\documentclass[11pt,a4paper]{article}
\usepackage[hyperref]{eacl2021}
\usepackage{times}
\usepackage{latexsym}

\usepackage{microtype}

\usepackage{multirow}

\usepackage{amsmath}
\usepackage{graphicx}
\usepackage{verbatim}
\usepackage{booktabs}

\usepackage{pgfplots}
\pgfplotsset{compat=1.16}

\aclfinalcopy 

\title{Low-Resource Language Modelling of South African Languages}

\author{Stuart Mesham \hspace{10pt}
  Luc Hayward \hspace{10pt}
Jared Shapiro \hspace{10pt}
Jan Buys \\
  Department of Computer Science \\
  University of Cape Town, South Africa \\
  \texttt{\{MSHSTU001,HYWLUC001,SHPJAR002\}@myuct.ac.za,jbuys@cs.uct.ac.za} }  
\date{}

\begin{document}
\maketitle
\begin{abstract}

Language models are the foundation of current neural network-based models for natural language understanding and generation.
However, research on the intrinsic performance of language models on African languages has been extremely limited, which is made more challenging by the lack of large or standardised training and evaluation sets that exist for English and other high-resource languages. 
In this paper, we evaluate the performance of open-vocabulary language models on low-resource South African languages, using byte-pair encoding to handle the rich morphology of these languages. 
We evaluate different variants of $n$-gram models, feedforward neural networks, recurrent neural networks (RNNs), and Transformers on small-scale datasets. 
Overall, well-regularized RNNs give the best performance across two isiZulu and one Sepedi datasets. 
Multilingual training further improve performance on these datasets.
We hope that this research will open new avenues for research into multilingual and low-resource language modelling for African languages.  

\end{abstract}

\section{Introduction}

Language modelling has applications in many areas of NLP including machine translation, information retrieval, voice recognition and question answering \cite{Wu2016, Franz2002, Chavula2016, Ndaba2016, Kumar2016}. Improvements in language modelling have resulted in improved model performance in the above tasks, making language modelling a valuable area of study. 
High resource languages have enjoyed substantial improvements in language modelling performance in recent years due to large neural models such as GPT-2, BERT and XLNet \cite{Radford2019, Devlin2019, Yang2019}.
However, most African languages are low-resource, and the limited availability of high-quality training data makes training large language models challenging.

\begin{figure}
    \centering
    \includegraphics[scale=0.5]{
    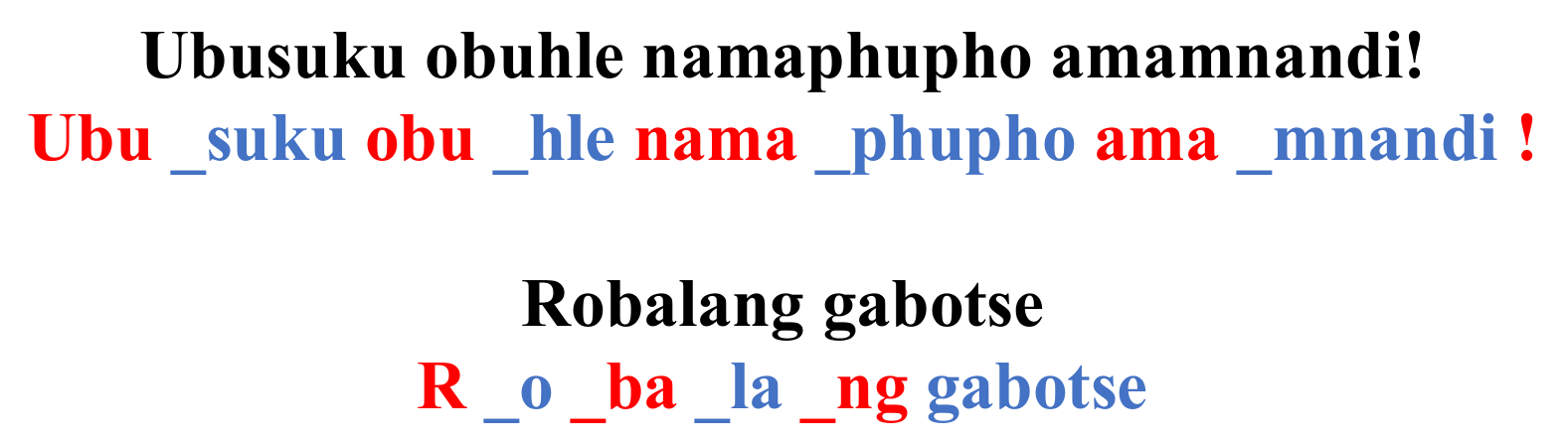}
    \caption{Example sentences and their BPE tokenizations in isiZulu (top) and Sepedi (bottom). The tokenizers use BPE vocabulary sizes of 8000 and 2000 respectively.}
    \label{fig:tokenization_example}
\end{figure}

In this paper we focus on South African Benue-Congo languages,
which are more resourced than most other Benue-Congo languages, but still clearly low-resourced.\footnote{The Benue-Congo languages is a subdivision of the Niger-Congo language family. Most Benue-Congo languages are part of what linguists refer to as the Bantu or Bantoid sub-families.}
The two groups of South African languages with the largest number of total speakers are the Nguni and Sotho-Tswana groups of closely-related languages. In South Africa these languages represent 43.3\% and 24.7\% of speakers respectively \cite{Census2011}. 
In our data sources, the isiZulu and Sepedi  languages had the largest amounts of text available, respectively, within these language groups.

In addition to the lack of large amounts of high quality data, Benue-Congo languages are  typologically\footnote{Typology refers to the linguistic properties and characterization of a language.} very different from the Indo-European languages most widely studied for language modelling. 
Even in large multilingual studies, African languages are usually underrepresented if included at all. 
Benue-Congo languages are agglutinative and morphologically rich \cite{pretorius-bosch-2009-exploiting}:
Most words are made up by combination of smaller morphological units, grammatical relations (such as subject or object) are indicated by changes in the words rather than the relative position of words in the sentence, and all nouns belong to one of a large number of noun classes which governs the choice of many morphemes. 
This leads to potentially very large and sparse word-level vocabulary, even though individual morphemes or sub-words many be more frequent in a corpus (as they are used in many different words).

This paper examines the application of n-gram models \cite{Chen1999}, Feed-forward neural networks (FFNNs) \cite{Bengio2003}, Recurrent neural networks including Long Short Term Memory  \cite[LSTMs;][]{Hochreiter1997} and Transformer \cite{Vaswani2017} models on isiZulu and Sepedi.
We use byte pair encoding \cite[BPE;][]{Sennrich2016} to control the vocabulary size and to enable open-vocabulary language modelling (see Figure \ref{fig:tokenization_example}), making the choice of vocabulary size a hyperparameter of the models. 

Our results show that the relative performance of the different model classes is similar to what have been found in previous work on small-scale language modelling in English and other languages.
Well-regularized RNNs, the AWD-LSTM \cite{Merity2017} and QRNN \cite{Bradbury2017}, have the best overall performance, outperforming the Transformer.
The $n$-gram, FFNN and baseline LSTM models performed worse across all datasets.
We also perform an evaluation of multilingual training, showing that training on text from multiple related languages improves performance without any modifications to the model architecture. 
The benefits can be seen using text from either the same language group or a different but related language group, despite orthographic differences. Code and trained models can be found at \url{https://github.com/StuartMesham/low\_resource\_lm}.

\section{Background}
\label{sec:background}

A language model assigns a probability $P(W_1^n)$ to a sequence of $n$ words $W_1^n=w_1,...,w_n$. 
The probability is usually decomposed using the chain rule to predict the words one at a time (from left to right) by assigning a probability to each word for following the given context~\cite{Daniel2020}: 
\begin{align}
P(W_1^n) &= \prod_{k=2}^n{P(w_k|W_1^{k-1})}.
\end{align}

\subsection{Sub-word Tokenization}
\label{sec:bpe}
Language models traditionally estimate the next word probability as a distribution over a fixed vocabulary, where the input text has been tokenized into words, and all words outside the vocabulary replaced with a special \emph{unknown} token.
South African Benue-Congo languages are highly agglutinative, making whole-word tokenization sub-optimal for language modelling due to potentially large vocabulary sizes and subsequent data sparsity. In contrast, character-level tokenization requires the model to learn to model very long sequences. To better represent the structure of the languages, we use byte-pair encoding \cite{Gage1994,Sennrich2016} 
to break words into sub-word units based on their frequency. 
Language modelling with BPE has previously been shown to perform competitively for open-vocabulary language modelling \cite{Mielke}.

Byte-pair encoding is a compression algorithm which has been adapted for sub-word tokenization. The algorithm starts with character-level tokens and finds pairs of adjacent tokens which occur most frequently. These token pairs are replaced with single tokens containing the concatenation of the characters in each token. This process is repeated until a desired vocabulary size is reached \cite{Sennrich2016}.
To ensure fair model evaluation, we train BPE tokenizers using only the training sets.
Example BPE tokenizations in isiZulu and Sepedi are shown in Figure~\ref{fig:tokenization_example}.

\subsection{Evaluation}

The quality of a language model can be evaluated either extrinsically or intrinsically. Extrinsic evaluation measures a model's usefulness in some downstream task such as speech recognition or machine translation whereas intrinsic evaluation uses statistical measures to assess a model's quality. 
In this paper we focus on intrinsic evaluation metrics related to cross-entropy and perplexity.

In information theory, entropy represents the average number of units of information produced per observation \cite{Shannon1948}.
The cross-entropy of a language model on a given sample of text $W_1^n$ is estimated as 
\begin{equation}\label{eq:cross_entropy}
H(W_1^n)=-\frac{1}{n}\log_2{P(W_1^n)},
\end{equation}
with the units of information being bits due to the log base 2 \cite{Daniel2020}. 
The more accurately the model approximates the true distribution of the language, the lower the cross-entropy.
Language models with a fixed vocabulary are usually evaluated based on perplexity, which is computed as $2^{H(W_1^n)}$.
However, closed-vocabulary language models have to set the size of the vocabulary and treat all other words as unknown. 
Consequently, perplexity cannot be compared directly across models with different vocabularies.

In this paper we are studying open-vocabulary models, and we want the choice of tokenization and vocabulary to be a modelling choice. 
This necessitates an evaluation metric which is independent of the tokenization.

As evaluation metric we use bits per character (BPC), a measure of cross-entropy which is normalised by the character length of the text and is therefore independent of the tokenization. 
The BPC of a model on a test set $W_1^n$ is calculated as
\begin{equation}
\mathrm{BPC}(W_1^n) = \frac{n}{c} H(W_1^n),
\end{equation}
where the text consists of $c$ characters.

\subsection{Models}

\subsubsection{$n$-gram Models}
$n$-gram language models make the Markov assumption of restricting the context for predicting the next word to the last $n-1$ words \cite{Daniel2020}.
Traditional $n$-grams are based on various smoothing methods, of which modified Knesser-Ney smoothing has been shown to lead to the best performance in general \cite{Kneser1995,Chen1999}.
Sparsity increases as the $n$-gram size increases, which leads to practical limits on the size of $n$ that is used. 

\subsubsection{Feedforward Neural Networks}

The first neural network-based language models were based on feedforward neural networks (FFNNs), which also make the Markov assumption, and are therefore effectively neuralized $n$-gram models \cite{Bengio2003}.
One of the key advantages of neural language models over $n$-grams is that word embeddings allow them to generalise better, as words with similar meanings or grammatical functions will have similar embeddings \cite{Mikolov2013b}.

The first layer of an FFNN takes the concatenation of the context word embeddings as input.
The embedding layer is learned jointly with the rest of the model and weight-tied to the output layer, following standard practice in RNN-based language modelling.
We use a rectified linear unit as non-linearity.

\subsubsection{LSTMs}

LSTMs \cite{Hochreiter1997} are a widely used variant of the standard RNN architecture allowing for longer term dependencies to be modelled more effectively by using a number of gates along with a memory vector in the recurrent cell. The gates and the memory vector enable information to pass more effectively across time steps. 
We use a Basic-LSTM model as a baseline for the more complex AWD-LSTM and QRNN models (see below).

This model is regularized using dropout, which temporarily hides a random subset of neurons during each training step \cite{Srivastava2014}.
This adds noise and prevents the model from being overly reliant on any particular neuron. 
However, dropout in RNN models cannot be applied between time steps on the recurrent connection as it inhibits the model's ability to retain long term dependencies, so the standard approach is to apply dropout only on the input and output connections \cite{Zaremba2015}. 
The Basic-LSTM baseline does not use the more complex regularization and optimization techniques used by the other models.

\subsubsection{AWD-LSTM}

The AWD-LSTM model \cite{Merity2018} is used widely for language modelling and forms the bases of the current state-of-the-art language modelling on small English datasets without dynamic evaluation \cite{takase-etal-2018-direct}.
In order to enable a fair comparison across models we are not using a continuous cache pointer
\cite{Grave2017} or dynamic evaluation.

The AWD-LSTM uses a number of improved regularization and optimization techniques.
Regularization is particularly important in low-resource settings. 
DropConnect \cite{wan2013regularization} 
is a form of dropout on the hidden-to-hidden weights.\footnote{This method is particularly useful as it is applied once to the weight matrices before the forward and backward pass, allowing the use of black box RNN implementations such as NVIDIA's cuDNN LSTM which can be many times faster due to hardware optimisations \cite{Merity2018}.} 
Variational dropout \cite{Gal} generates a dropout mask once which is then used over the entire forward and backward pass, rather than resampling at every timestep. 
The AWD-LSTM model uses a combination of DropConnect for the hidden-to-hidden transitions within the LSTM and variational dropout over the inputs and outputs. 
Other tehchniques used include using variable length backpropagation sequences, word dropout (masking entire word embeddings),
and L1 and L2 regularisation. 

\subsubsection{Quasi-Recurrent Neural Networks}

Quasi-Recurrent Neural Networks (QRNNs) \cite{Bradbury2017} is a modification of RNNs that parallizes parts of the RNN computation and obtained similar or even slightly better performance than the AWD-LSTM on some English datasets \cite{Merity2018a}.
The QRNN applies convolutional layers on the input, followed by an recurrent pooling function resembling LSTM gating. 
This significantly increases training speed compared to LSTMs of similar sizes.

\subsubsection{Transformers}

The Transformer \cite{Vaswani2017} presents another approach to speeding up sequential processing over RNNs by relying entirely on attention mechanisms \cite{Bahdanau2015} instead of recurrent connections for propagating information across time steps. 
An attention mechanism can process all the input embeddings for a (fixed-length) sequence simultaneously and selectively weight certain features based on a learned function. 

The original Transformer model was used for translation and has an encoder-decoder structure \cite{Vaswani2017}. For the task of language modelling, only the decoder architecture is used \cite{Liu2018}. We follow the architecture used by GPT-2 \cite{Radford2019}.
A learned positional embedding is added to each input token embedding. 
Multiple layers, each including an attention and a feedforward sub-layer, are stacked to create the larger model that can propagate information more efficiently across time steps.  
In each attention sub-layer multiple attention mechanisms are used to extract features; this strategy is termed multi-headed self-attention.
Finally, a residual connection and layer normalisation is applied over each sub-layer. 
To regularise the Transformer models we use dropout on all weights of the model.

\section{Experimental Setup}

\subsection{Datasets}

We focus on language modelling for isiZulu and Sepedi, but we processed data for all 11 non-European official South African languages, and use the other languages' data for multilingual training (Section \ref{sec:multilingual}).  
We use two dataset sources:

\textbf{NCHLT:} 
We use the corpora from the National Centre for Human Language Technology (NCHLT) Text project \cite{Eiselen2014} made available by the South African Centre for Digital Language Resources (SADiLaR).\footnote{Datasets are available at \url{https://repo.sadilar.org/handle/20.500.12185/7}} 
Monolingual text corpora are available for all 11 of South Africa's official languages. 
We processed the corpora for the Nguni languages
(isiZulu, Siswati, isiNdebele and isiXhosa) and the Sotho-Tswana languages (Sesotho, Sepedi, Setswana), as well as Xitsonga and Tshivenda, the other two Benue-Congo languages.
A significant proportion of these texts were scraped from governmental websites. The corpora range in size from 1 to 3 million tokens. 
Sepedi and isiZulu have the largest datasets in their respective language groups. 
    
\textbf{Isolezwe:} 
News articles from the isiZulu Isolezwe newspaper, one of the largest daily African language newspapers in South Africa, have been scraped and consolidated by the Newstools initiative.\footnote{Available at \url{https://github.com/newstools}}
This is the largest publicly-available newspaper corpus among the languages we are considering that we are aware of.
The dataset has a similar size to the NCHLT isiZulu corpus but provides a second evaluation domain. 

\begin{table}[t]
\centering
\begin{tabular}{lcc}
\toprule
\textbf{Corpus} & \multicolumn{2}{c}{\textbf{Words}} \\ 
 &  Training & Valid/Test \\
\midrule
NCHLT (isiZulu) & 978.6 & 122.3 \\
Isolezwe (isiZulu) & 940.2 & 117.5 \\
NCHLT (Sepedi) & 1357.3 & 169.7 \\
\bottomrule
\end{tabular}
\caption{Dataset sizes, reported in thousands of words, after preprocessing. The validation and test sets of each corpus are approximately equal in size.
}
\label{tab:datasets}
\end{table}

We performed a number of data proprocessing and normalization steps.
We removed instances of English, HTML and Javascript lines, and other repetitive or erroneous data, as these would not naturally be found in general language.
Each dataset was split into a training, validation and test set using an 80\% / 10\% /10\% split. 
The splits were done using sequential blocks 
to preserve the order of the sentences.
Table \ref{tab:datasets} compares the dataset sizes. 

\subsection{Model Implementation and Optimization}

The BPE preprocessing for all models uses the HuggingFace tokenizers library.\footnote{\url{https://github.com/huggingface/tokenizers}}

\subsubsection{$n$-gram Models}

We use an $n$-gram language model with
modified Kneser-Ney \cite{Chen1999} smoothing, as implemented in KenLM.\footnote{Available at \url{https://github.com/kpu/kenlm}}
We tuned the models by testing BPE vocabulary sizes ranging from 100 to 10000 and $n$-gram orders from 2 to 6. The isiZulu and Sepedi models performed best with BPE vocabulary sizes of 500 and 2000 respectively. For all datasets, an $n$-gram order of 6 yielded the best performance. 

\subsubsection{Feedforward Neural Networks}

We implemented a feed-forward neural network (FFNN) language model so that it can be trained in a similar manner to RNN and Transformer language models.
The training data is divided into chunks of 64 tokens and batched to enable parallel processing. 
We follow the optimization and regularization setup of the FFNN baseline used by  \citet{chiu-rush-2020-scaling}.
We use of a learning rate decay schedule where the 
learning rate is multiplied by 0.25 after each epoch if the validation loss does not improve. 
The models were trained for 50 epochs with a batch size of 32 and an AdamW weight decay of 0.01. Both word embeddings and hidden layers had a size of 500.

Using grid search, we evaluated BPE vocabulary sizes 1000 and 2000 to 10 000 with an interval size of 2000, 
$n$-gram orders \{2, 4, 6\},
word embedding and hidden layer sizes in the range \{500, 2500\} with an interval of 250, 
dropout rates of \{0.3, 0.5\}
and \{2, 4, 6\} hidden layers.

For both NCHLT isiZulu and NCHLT Sepedi a BPE vocabulary size of 8000 yielded the best performance, and on Isolezwe 10 000 performed best. For both Isolezwe and NCHLT isiZulu, an $n$-gram order of 2 performed best and for NCHLT Sepedi an order of 4.
We were unable to find a fully satisfactory explanation of why the FFNN did not perform better with a higher $n$-gram orders.

\subsubsection{LSTMs}

We use the PyTorch implementation of the AWD-LSTM
\cite{Merity2018}.\footnote{Available at  \url{https://github.com/salesforce/awd-lstm-lm}}
We took the hyperparameters of \citet{Merity2018}
on the word-level WikiText 2 dataset as the starting point for tuning our models, as its size is comparable to our dataset.
We performed a partial grid search over the embedding size \{400, 800\}, hidden layer size \{1150, 1200, 1550\}, number of layers \{1, 2, 3, 4\}, learning rate \{5, 10, 30\}, batch size \{40, 80\}, vocab size \{2500, 5000, 7500, 10000\} as well as dropout rate \{0 - 0.7\} and weight drop \{0 - 0.5\} (both in increments of 0.1) and L1/L2 regularisation values \{0, 1, 2\}.
Model development was primarily done on the isiZulu NCHLT corpus. 
Most improvement came from increasing the total model size by either increasing the number of hidden layers or increasing the input embedding size. 
Changing the BPE vocabulary size did not have a significant effect on performance.
The Basic LSTM was tuned similarly, excluding the regularization techniques it does not implement. 

\subsubsection{QRNN}

The QRNN is also implemented in the AWD-LSTM packages.
We tuned the embedding size, vocabulary size, number of hidden layers and batch size, using similar ranges as for the AWD-LSTM. 
The best QRNNs used an embedding size of 800, hidden layer sizes of 1550, and 4 hidden layers. 

\begin{figure}
    \centering
    \includegraphics[scale=0.9]{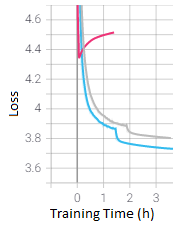}
    \caption{The validation loss of the basic LSTM (pink), AWD-LSTM (grey) and QRNN (blue) while training on the NCHLT isiZulu dataset. 
} 
    \label{fig:nzawd}
\end{figure}

Figure \ref{fig:nzawd} shows how the validation loss changes while the RNN-based models (Basic LSTM, AWD-LSTM, and QRNN) train on the NCHLT isiZulu corpus.
The plot shows how the QRNN's loss decreases faster than that of the AWD-LSTM time. 
The Basic LSTM initially trains faster, but then overfits drastically.

\subsubsection{Transformers}

We used the GPT-2 \cite{Radford2019} PyTorch implementation provided by the open-source HuggingFace transformers library.\footnote{\url{https://huggingface.co/transformers/}} 
The training data was fed to the model in blocks of 128 consecutive tokens with a batch size of 32, 
created using a sliding window over the training data with a stride of 16 tokens. 
Model evaluation was performed using an input block size of 128 with a stride of 64.

For hyper-parameter tuning, models were trained for up to 200k steps, with evaluation on a validation set every 5k steps. Training was stopped early if the validation loss did not decrease after any four successive evaluations. 
The model and vocabulary sizes were tuned first with little regularization to ensure that the models had enough capacity to overfit the data. 
Increasing amounts of regularization were then applied until the model no longer overfit the data. 

We used 8 hidden layers and 8 attention heads. 
Preliminary experiments showed that the model was relatively insensitive to the number of hidden layers and the number of attention heads.
We used an initial learning rate of  $10^{-4}$ with a learning rate schedule that linearly decreases to 0 over the course of the training.
Across all 3 corpora, the best performing models had a hidden layer size of 256, a dropout probability of 0.3 and a weight decay of 0.2. The isiZulu and Sepedi models performed best with BPE vocabulary sizes of 8000 and 2000 respectively.

\section{Results and Discussion}

\begin{table*}[t]
\centering
\begin{tabular}{lccccccccc}
\toprule
 & \multicolumn{3}{c}{\textbf{NCHLT (isiZulu)}} &
\multicolumn{3}{c}{\textbf{Isolezwe (isiZulu)}} & \multicolumn{3}{c}{\textbf{NCHLT (Sepedi)}}
\\
\cmidrule(lr){2-4}\cmidrule(lr){5-7}\cmidrule(lr){8-10}
\textbf{Model} & Params & Vocab & BPC & Params & Vocab & BPC & Params & Vocab & BPC \\
\midrule
$n$-gram   & 7.5M & 500  & 1.588 & 6.9M & 500 & 1.544 & 5.7M &  2000 & 1.656 \\
FFNN       & 4.7M & 8000 & 1.572 & 5.7M & 1000 & 1.532 & 5.1M & 8000 & 1.723 \\ 
\midrule
Basic-LSTM & 3.3M & 5000 & 1.548 & 3.3M & 5000 & 1.677 & 3.3M & 5000 & 1.625 \\
AWD-LSTM   & 29.8M & 5000 & 1.325 & 29.8M & 5000 & \textbf{1.259} & 29.8M & 5000 & 1.421 \\
QRNN       & 29.5M & 10000& \textbf{1.323} & 29.5M & 10000& 1.264 & 29.5M & 5000 & \textbf{1.421} \\
\midrule
Transformer & 8.6M & 8000 & 1.391 & 8.6M & 8000 & 1.320 & 7.1M & 2000 & 1.495 \\  
\bottomrule     
\end{tabular}
\caption{Language modelling results on the isiZulu and Sepedi corpora, reported as bits-per-character (BPC). The BPE vocabulary size and number of parameters of each model are also reported.}
\label{table:all-test-results}
\end{table*}

\subsection{Results}

All the test set results are given in Table \ref{table:all-test-results}.
The $n$-gram and FFNN language models performed fairly similarly to each other across the datasets and languages, even though the FFNNs used smaller $n$-gram orders. 
On the isiZulu datasets, the FFNN performed slightly better than the $n$-gram models, while on the Sepedi dataset the $n$-gram model performed better. 
On all datasets, we found that the $n$-gram models tended to perform better with smaller BPE vocabulary sizes, whereas the FFNN models performed better with larger vocabulary sizes.

The performance of the AWD-LSTM and QRNN models was closely matched (within 0.005 BPC) across all datasets with the QRNN slightly outperforming on the two NCHLT datasets, and the AWD-LSTM ahead on the Isolezwe dataset.
The basic LSTM under-performed the others substantially, with performance closer to, or even worse than, that of the $n$-gram and FFNN models.

The transformer models achieved competitive performance on all datasets, but were outperformed by the QRNN and AWD-LSTM.
We hypothesize that the main reason is that these models used more sophisticated regularization techniques that our Transformer implementation did not use. 
Additionally, the RNNs had more parameters, but the Transformer's performance did not improve with more parameters in our experiments.

\subsection{Discussion}

\begin{figure}
    \centering
    \includegraphics[scale=0.36]{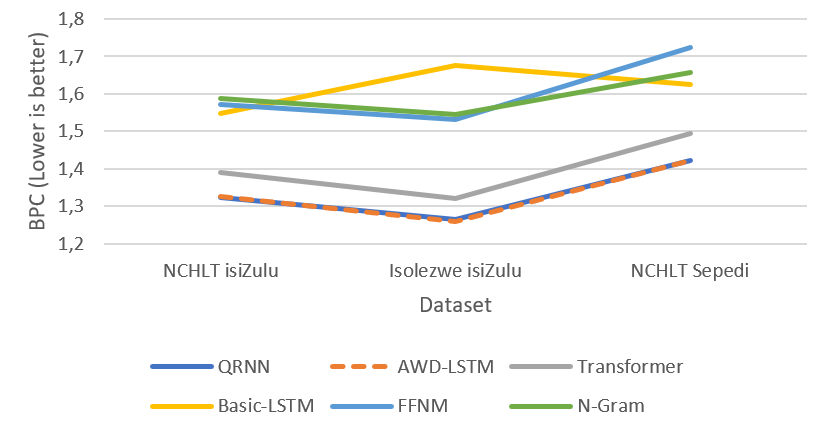}
    \caption{
    Test set results, plotted to show the relative performance of the models on each of the three datasets (lower is better). The AWD-LSTM and QRNN consistently outperform the other models while within close margin of each other, followed by the Transformer, while the $n$-gram, FFNN and Basic-LSTM perform substantially worse.}
    \label{fig:relative}
\end{figure}

The results show that the relative performance of the models is similar to language modelling results previously reported on widely used PTB and WikiText2 English datasets \cite{Merity2017}, which are comparable in size to our corpora.
Regarding the performance of the Transformer, it has been reported that a modified Transformer architecture with segment-level recurrence can obtain similar results to the AWD-LSTM when fine-tuned using the same sophisticated regularization techniques \cite{Dai2019}, but other researchers have struggled to reproduce these results independently.\footnote{\url{https://twitter.com/srush_nlp/status/1245825437240102913}}

We found that the relative performance of the language models was similar across the three datasets (Figure \ref{fig:relative}).
This supports the hypothesis that the same models would likely perform well across all the languages in the Nguni and Sotho-Tswana language groups.
The AWD-LSTM and QRNN models were consistently close in performance, followed by the transformer model across all datasets. 
The remaining $n$-gram, FFNN and Basic-LSTM models had different relative performances on the datasets with no consistent pattern, although the $n$-gram and FFNN are closer to each other. 
The poor performance of the $n$-gram and FFNN models represents a trade-off between training time and model performance. If training time was a factor, reduced performance could be accepted in order to produce models more quickly. The $n$-gram models are also much faster when queried in downstream applications.

\section{Multilingual Models}
\label{sec:multilingual}

\begin{figure}[t]
\centering
\begin{tikzpicture} 
    \begin{axis}[
    ybar, 
    ymin=1.25,
    ymax=1.54,
    symbolic x coords={isiZulu, Sepedi},
    xtick=data,
    ylabel={BPC},
    bar width=0.6cm,
    x=3cm, 
    nodes near coords={\pgfmathprintnumber[fixed zerofill, precision=3]{\pgfplotspointmeta}}, 
    nodes near coords style={font=\small}, 
    enlarge x limits={abs=1.8cm}, 
    legend style={at={(0.5,-0.15)}, anchor=north,legend columns=2}, 
    ]
        \addplot coordinates {(isiZulu, 1.391) (Sepedi, 1.495)};
        \addplot coordinates {(isiZulu, 1.334) (Sepedi, 1.447)};
        \addplot coordinates {(isiZulu, 1.331) (Sepedi, 1.477)};
        \addplot coordinates {(isiZulu, 1.298) (Sepedi, 1.416)};
        \legend{Monolingual, Same Group, Different Group, All Languages}
    \end{axis}
\end{tikzpicture}
\caption{
Multilingual language modelling results, reported as bits-per-character (BPC), evaluated on the isiZulu and Sepedi test sets. 
Models were trained on the target language (Monolingual), and additionally also on multiple languages in the same language group (Nguni and Sotho-Tswana, respectively), languages from the other language group, or on text from all 9  non-European official South African languages.
}
\label{fig:multilingual_pretraining_graph}
\end{figure}
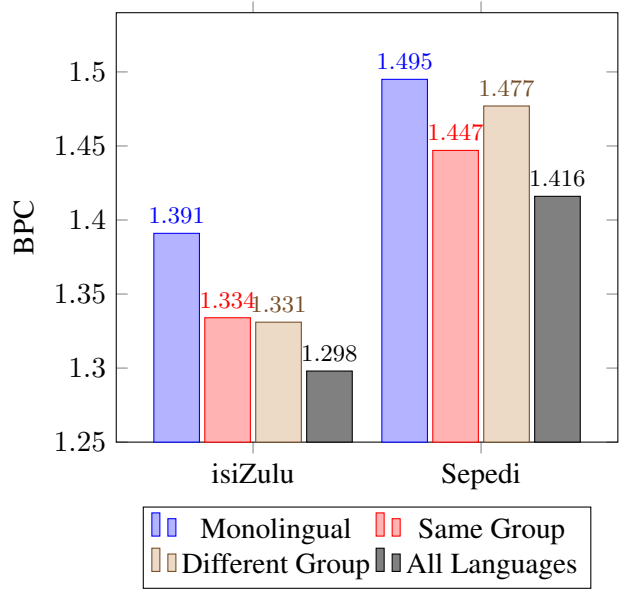

As an additional experiment, we investigate 
the potential for multilingual language modelling by concatenating training data from multiple languages and evaluating on the same target languages as before. For practical reasons, we only train Transformer models for this experiment.
We use the NCHLT corpora as they provide text in the same domain across all South African languages.

We train models in a number of different settings.
In particular, we were interested in comparing the effect of training on additional languages from the same language group (isiZulu: all Nguni languages; Sepedi: all Sotho-Tswana languages) compared to training on languages from the other language group (isiZulu: Sotho-Tswana languages; Sepedi: Nguni languages).
Finally, we also evaluated a model trained on all 9 Benue-Congo South African languages in the NCHLT corpus. Model hyper-parameters were tuned separately in each instance.

The results are shown in Figure~\ref{fig:multilingual_pretraining_graph}.
For both target languages, concatenating training data from other Benue-Congo languages improves performance. 
In general, training on more languages improves performance regardless of the language group.
In the case of Sepedi as target language, concatenating the other Sotho-Tswana languages yields a greater performance improvement than concatenating Nguni languages. 
On the other hand, for isiZulu the results of including additional data from the same or the other langauge family were similar. For both isiZulu and Sepedi models, the best performance is obtained by concatenating data from all languages. 
We hypothesize that transfer may be more effective from disjunctively written languages (Sotho-Tswana) to conjunctively written languages (Nguni) than the other way around, but this needs to be investigated further.
Our results suggest that the use of data from multiple languages is a promising future direction for modelling South African languages.

\section{Conclusions}

The experiments conducted in this paper demonstrated that improved regularization techniques and model architectures developed on relatively small English datasets also improves language modelling performance when applied to African languages such as isiZulu and Sepedi. 
The AWD-LSTM and QRNN performed notably better than the other models. 
As expected, $n$-grams and FFNNs, as well as the Basic LSTM, under-performed the more advanced models. 
However, the stronger models are computationally more expensive. 
Our results suggest that further improvements in RNN- and Transformer-based language modelling would likely be directly applicable to low-resource African languages. 
Additionally, we showed that BPE is an effective method for open vocabulary language modelling across multiple models, effectively accounting for the large (word-level) vocabulary sizes of agglutinative African Languages.
Finally, we showed that multilingual language modelling is a promising direction for future research, as many African languages occur in groups of closely related languages which might benefit from such an approach.

\section*{Acknowledgments}

This work is based on research supported in part by the National Research Foundation of South Africa (Grant Number: 129850) 
and the South African Centre for High Performance Computing.

\bibliography{anthology,eacl2021}
\bibliographystyle{acl_natbib}

\end{document}